\documentclass[11pt]{article}

\usepackage{arxiv}

\usepackage{amsmath,amsfonts,bm}

\def\eqref#1{equation~\ref{#1}}

\def\1{\bm{1}}

\DeclareMathAlphabet{\mathsfit}{\encodingdefault}{\sfdefault}{m}{sl}
\SetMathAlphabet{\mathsfit}{bold}{\encodingdefault}{\sfdefault}{bx}{n}

\usepackage[round]{natbib}
\usepackage{capt-of}
\usepackage{booktabs,multirow,makecell,graphicx,xcolor}
\usepackage{url}
\usepackage[font=small,labelfont=bf,skip=6pt]{caption}
\definecolor{uclablue}{rgb}{0.15, 0.45, 0.68}
\usepackage[
    pagebackref,
    breaklinks,
    citecolor=uclablue,
    linkcolor=uclablue,
    urlcolor=uclablue,
    colorlinks=true,
]{hyperref}

\newcommand{\second}[1]{\textcolor{blue!70!black}{\textbf{#1}}}
\newcommand{\third}[1]{\textcolor{green!50!black}{\textbf{#1}}}

\makeatletter
\g@addto@macro\@maketitle{%
  \begin{center}
    \includegraphics[width=\linewidth]{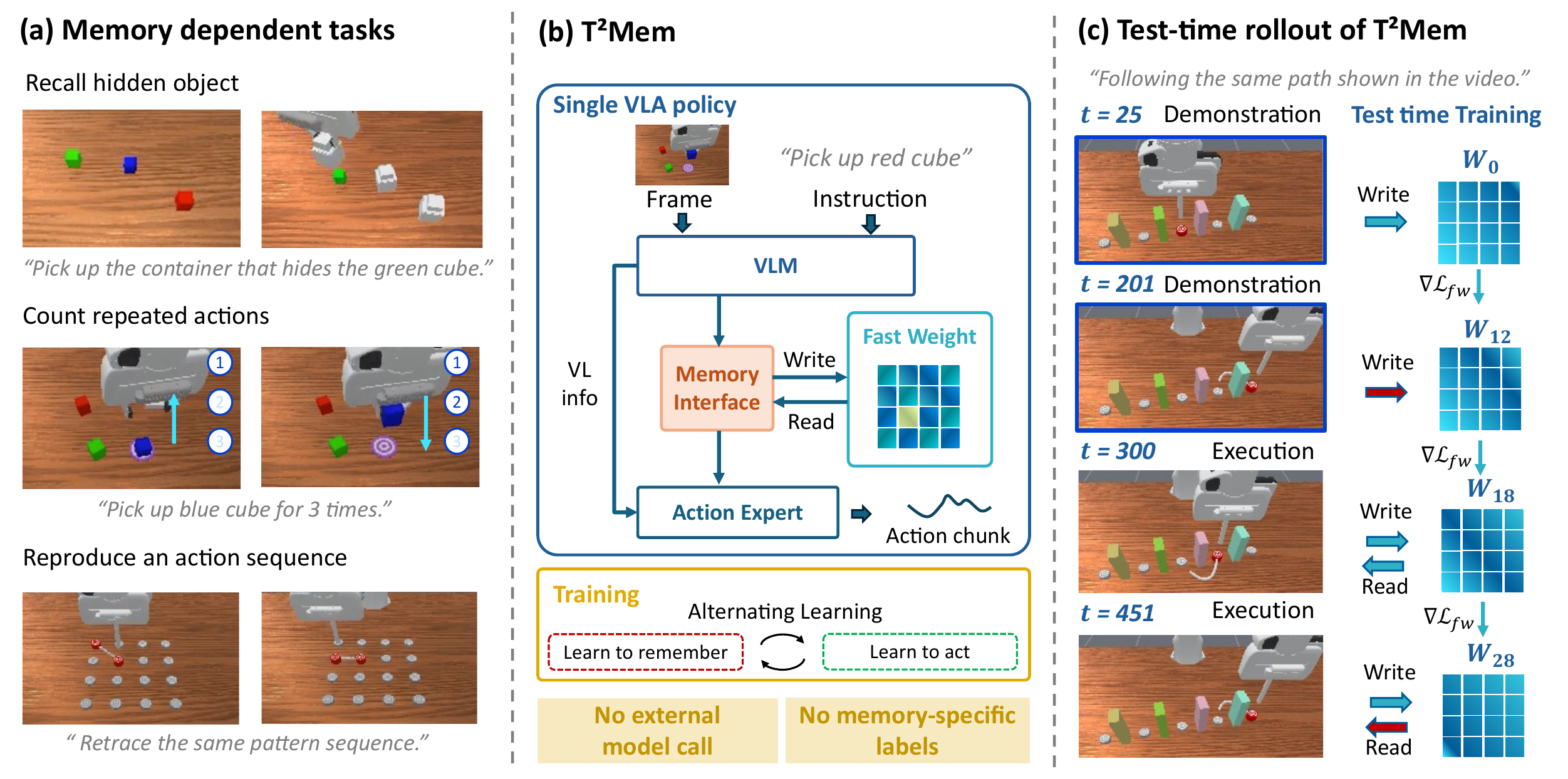}
    \captionof{figure}{\textbf{Test time memory (T²Mem): learning to remember for memory-dependent robot manipulation.}
(a) Tasks such as recalling hidden objects, counting repeated actions, and reproducing action sequences require information beyond the current observation.
(b) T²Mem integrates fast-weight memory into a single VLA policy through an observation-grounded interface, making vision--language history information available to the action expert. Alternating memory--policy learning develops the ability to retain and use history, without external model calls or memory-specific annotations.
(c) During a rollout, the policy uses test-time training (TTT) to encode observation history in fast weights through self-supervised updates. It retrieves this information to guide actions when relevant evidence is no longer visible.}
    \label{fig:teaser}
  \end{center}
  \vspace{0.5em}%
}
\makeatother

\title{T²Mem: Learning Test-Time Memory for Robotics}

\author{%
  \textbf{Yize Liu}$^{1}$ \quad
  \textbf{Huang Huang}$^{1}$ \quad
  \textbf{Yining Hong}$^{1}$ \quad
  \textbf{Zijian Du}$^{2}$ \quad
  \textbf{Zhi Cao}$^{3}$ \quad
  \textbf{Li Fei-Fei}$^{1}$ \quad
  \textbf{Jiajun Wu}$^{1}$ \\[0.6em]
  $^{1}$Stanford University \qquad
  $^{2}$NVIDIA \qquad
  $^{3}$University of Michigan, Ann Arbor \\[0.4em]
  {\small\texttt{yizeliu@stanford.edu}} \\[0.3em]
  \href{https://yzliu84.github.io/T2MEM-project/}{\textbf{\texttt{https://yzliu84.github.io/T2MEM-project/}}}
}

\renewcommand{\shorttitle}{T²Mem: Learning Test-Time Memory}
\date{}

\begin{document}
\newcommand{\tmem}{T²Mem}
\maketitle

\begin{abstract}
Memory-dependent robotic manipulation requires policies to use information that is no longer available in the current observation. Retaining history alone is insufficient: memory must preserve information that supports future actions. One challenge is whether a memory-free foundation model can learn to retain and use historical information from action demonstrations alone, without external memory support. We introduce T²MEM, a framework that develops this capability within a pretrained vision-language-action policy, without external reasoning models or memory-specific annotations. T²MEM uses test-time training to encode observation history into compact fast weights through online self-supervised updates, avoiding repeated processing of the full history. An observation-grounded interface extracts vision-language information for memory formation and supplies retrieved context to the action expert. Action supervision shapes what the memory learns to retain and use, while alternating memory–policy learning gives each component a fixed counterpart during optimization. Across 16 RoboMME tasks, T²MEM improves average success from 17.93\% to 56.83\% over the memory-free base policy and outperforms the recurrent-memory methods reported in the benchmark, while controlled profiling indicates at least 3$\times$ inference speedup over explicit methods. \href{https://yzliu84.github.io/T2MEM-project/}{Project website}.

\end{abstract}

\section{Introduction}
\label{sec:introduction}

Memory is not a verbatim record of the past, nor is it formed in isolation from the functions it serves: past experience is internally represented and later recovered in ways shaped by how it is ultimately used for future decisions~\citep{tulving1973encoding,morris1977levels,schacter1998constructive}. For robots, effective memory is therefore neither limited to the current observation nor retaining the full interaction history in a long context window, but instead lies in forming an internal memory state shaped by the decisions it must support --- for example, remembering which object was hidden to select the correct target after occlusion, or the order of a demonstrated sequence to reproduce the same sequence during execution. Memory and action are therefore reciprocal: memory guides future action, while action supervision shapes how memory is formed and used.

Such decision-relevant memory naturally belongs within the decision-making system itself, rather than being constructed as an explicit representation and then passed to the policy through separate storage, retrieval, or reasoning modules. Explicit memory introduces an intermediate bottleneck: information discarded or distorted before action prediction cannot be recovered downstream, and the memory representation is not necessarily shaped by the decisions it must support. This points to an implicit, end-to-end form of memory, where memory formation and use are jointly shaped by action, echoing cognitive accounts in which memory is reconstructive over internal representations~\citep{bartlett1932remembering}.
Test-time training (TTT) offers a natural computational mechanism for this view: instead of explicitly retaining history, past observations can be absorbed into fast weights through online self-supervised updates \citep{sun2020ttt,sun2024tttlayers}. These online updates can encode history. However, encoding history alone does not ensure that the resulting memory contains the information most useful for action. This motivates an internal memory mechanism whose online updates remain self-supervised, while its formation is shaped by action supervision toward downstream control.

We propose T²Mem, which integrates adaptive \textbf{T}est-\textbf{T}ime \textbf{MEM}ory directly into a pretrained VLA, making memory an intrinsic part of the policy. An \emph{observation-grounded memory interface} between the vision-language backbone and the action expert encodes observation history into fast weights through self-supervised updates and queries this evolving memory to support history-conditioned actions. Our \emph{alternating memory--policy learning} couples memory formation with its use in control: action supervision shapes how memory is formed and used, while alternating optimization gives memory and policy a fixed counterpart when learning to remember and to act, respectively. At deployment, the policy remains fixed while memory adapts from observations alone. The method requires no memory-specific annotations, auxiliary task-state targets, or external model calls, and learns from task demonstrations without an additional broad sequence-pretraining stage.

We evaluate T²MEM on RoboMME, which comprises 16 memory-dependent tasks across four categories \citep{dai2026robomme}. T²MEM improves average success from 17.93\% to 56.83\% over the memory-free base policy and outperforms the recurrent-memory methods reported in the benchmark. Memory interventions and efficiency analysis further examine whether the policy uses stored history and at what inference cost. Our central question is whether a given pretrained policy can learn to extract, retain, and use decision-relevant information through action supervision and online observation-based self-supervision, without larger external reasoning models or additional memory annotations. Beyond task success, we aim to make memory an intrinsic capability of a single policy. This approach could extend to more advanced foundation models, providing a foundational memory capability that complements explicit memory and higher-level reasoning for capable and efficient robotic control.

\section{Related Work}
\label{sec:related-work}

\paragraph{Memory in Robotics}
\label{sec:robot-memory}
Memory-augmented policies use history to address partial observability.
Active Neural SLAM maintains spatial maps~\citep{chaplot2020active}, while Scene Memory Transformer attends to stored observation embeddings~\citep{fang2019smt}.
Spatial maps primarily capture geometry; observation banks require managing retention and retrieval costs.
For manipulation, SAM2Act+ supports spatial recall through a memory bank~\citep{fang2025sam2act}, and MemoryVLA consolidates perceptual and semantic features~\citep{shi2026memoryvla}.
These approaches trade historical detail against storage and attention costs.
Recurrent methods instead compress history into persistent states.
RoboFlamingo uses a recurrent policy head~\citep{li2024roboflamingo}, while ReMem-VLA combines dual-level recurrent queries with auxiliary past-observation reconstruction~\citep{li2026rememvla}.
However, reconstructing past observations is not equivalent to retaining decision-relevant information.
MemoryBench and RoboMME evaluate whether historical information supports decisions beyond the current observation~\citep{fang2025sam2act,dai2026robomme}.

\paragraph{Test-Time Training}
\label{sec:ttt-related-work}
Fast-weight models store temporary associations in rapidly changing
parameters~\citep{ba2016fastweights}.
TTT adapts models through test-time self-supervision~\citep{sun2020ttt}.
Subsequent work introduces feature alignment~\citep{liu2021tttpp},
masked reconstruction~\citep{gandelsman2022tttmae}, and adaptation
over video streams~\citep{wang2025tttvideo}.
TTT layers encode history through online updates to a learned
memory model~\citep{sun2024tttlayers}, while Titans adds surprise-driven
updates, momentum, and forgetting~\citep{behrouz2025titans}.
Applying these mechanisms to robotics requires bridging visual
observations, semantic representations, and action generation:
memory must retain information in a form the policy can use.
RoboTTT integrates fast weights into robot policies, emphasizing
visuomotor context scaling and in-context adaptation~\citep{jiang2026robottt}.
However, longer context and memory-dependent decision making are
distinct objectives.
Tasks requiring semantic memory depend on selectively retaining
task-relevant cues and retrieving them when current observations
are insufficient, rather than merely extending the history
available to action prediction.

\section{T²Mem}

\label{sec:prelim}
\begin{figure}[!t]
  \centering
  \includegraphics[width=1\linewidth]{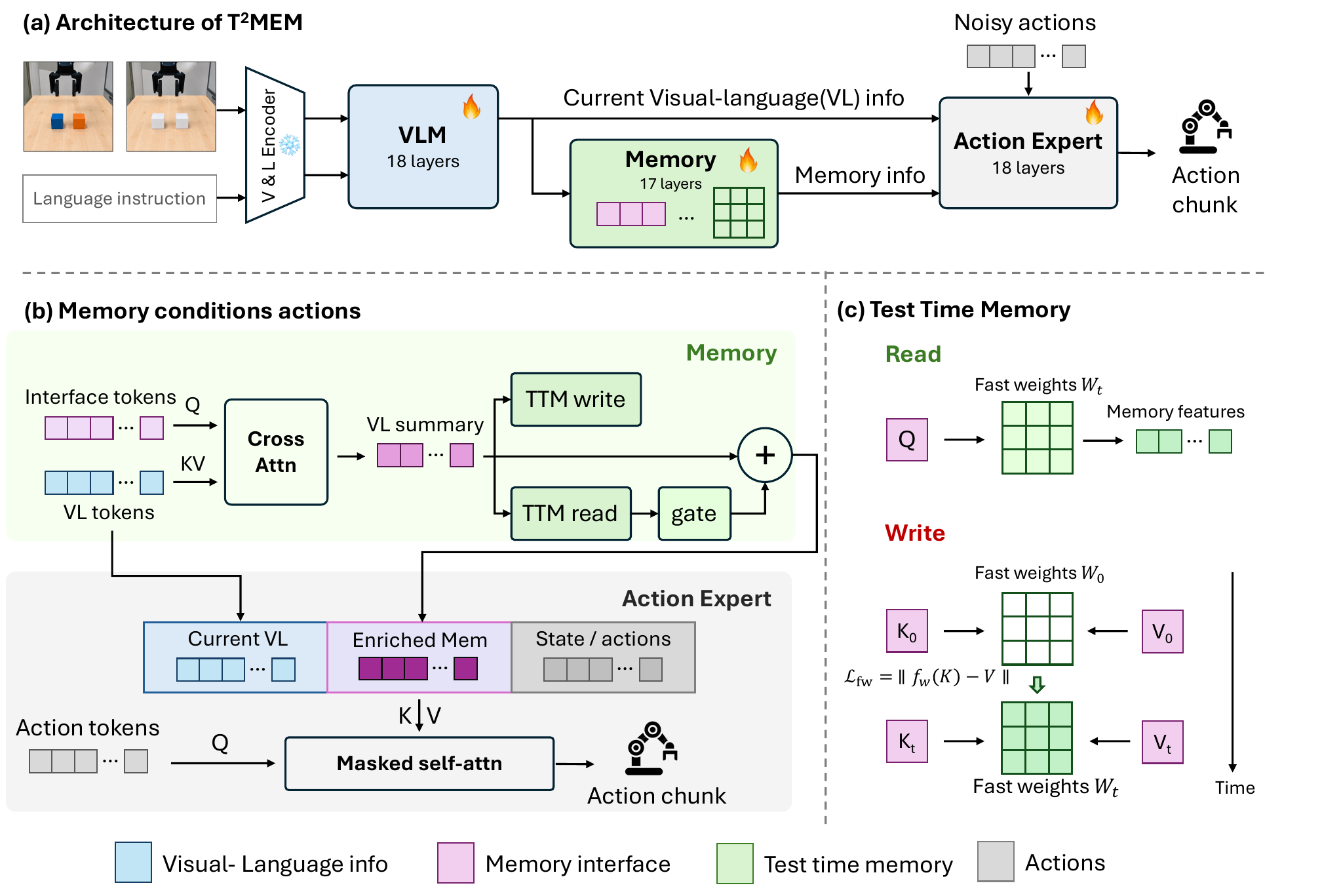}
  \caption{\textbf{Architecture of \tmem{}.}
  (a) Memory connects the VLM and AE alongside the direct vision-language pathway.
  (b) The observation-grounded interface extracts vision-language features, fuses memory readouts, and supplies attention context to the AE.
  (c) Self-supervised updates encode history in fast weights.}
  \label{fig:architecture}
\end{figure}

\tmem{} addresses memory-dependent tasks in which the current observation is insufficient for action selection (Figure ~\ref{fig:architecture}).
It connects a vision-language model (VLM) to an action expert (AE) through a parametric memory pathway.
The memory extracts vision-language features stores as semantic history through adaptive self-supervised updates, and provides historical context for action prediction.
No memory-specific annotations or auxiliary task-state targets are required.
Expert actions supervise the outer training objective; at deployment, memory updates use observations alone.

\subsection{Preliminaries}
\paragraph{Memory-Dependent Tasks}
\label{sec:memory-dependent-task}

Let $o_t$ denote the robot's visual and proprioceptive observation, $\ell$ a language instruction, and $h_t=(o_0,a_0,\ldots,a_{t-1},o_t)$ the interaction history.
Define $\Pi_{\mathrm{hist}}$ as the class of history-conditioned policies $\pi_t(a_t\mid h_t,\ell)$ and $\Pi_{\mathrm{current}}$ as the class of observation-conditioned policies $\pi_t(a_t\mid o_t,\ell)$.
Under a fixed task distribution and interaction budget, let $J(\pi)=\mathbb{E}_{\pi}[R(\tau)]$ denote expected task return for trajectory $\tau$, with $R$ optionally defined as the success indicator.
We call a task \emph{memory-dependent} if
\begin{equation}
  \sup_{\pi\in\Pi_{\mathrm{hist}}} J(\pi)
  >
  \sup_{\pi\in\Pi_{\mathrm{current}}} J(\pi),
  \label{eq:memory-dependence}
\end{equation}
so that the current observation alone is insufficient for optimal performance.
Both policy classes may depend on time $t$; the distinction concerns access to past observations and actions.

\paragraph{Test-Time Training}
\label{sec:ttt-preliminaries}

TTT updates a model at inference using a self-supervised objective constructed from its inputs.
In the fast-weight formulation, a neural model $f_{W_t}$ encodes episode history in its parameters $W_t$~\citep{sun2024tttlayers}.
Given an input representation $x_t$, slow parameters $\theta$ produce queries, keys, and values $(q_t,k_t,v_t)$.
A standard associative objective is
\begin{equation}
  \mathcal{L}_{\mathrm{mem}}(W;x_t)
  = \frac{1}{2}\left\|f_W(k_t)-v_t\right\|_2^2,
  \label{eq:ttt-inner-objective}
\end{equation}
with memory readout and update
\begin{equation}
  \begin{aligned}
    m_t &= f_{W_t}(q_t), \\
    W_{t+1} &= W_t - \eta_t
      \left.\nabla_W \mathcal{L}_{\mathrm{mem}}(W;x_t)\right|_{W=W_t}.
  \end{aligned}
  \label{eq:ttt-read-write}
\end{equation}
The action policy is conditioned on the readout, $\pi_{\theta}(a_t\mid o_t,\ell,m_t)$.
We adopt a read-before-write convention with inner-loop step size $\eta_t$; setting $\eta_t=0$ skips a write.
The slow parameters $\theta$ and initialization $W_0$ are learned through the outer action objective.
At deployment, $\theta$ remains fixed, while fast weights update without expert action labels and reset to $W_0$ at each episode boundary.

\subsection{Architecture of \texorpdfstring{\tmem{}}{T2MEM}}
\label{sec:architecture}

As shown in Fig.~\ref{fig:architecture}(a), \tmem{} builds on $\pi_{0.5}$~\citep{black2025pi05} and comprises a VLM backbone, a memory module, and an AE.
The memory sits between the VLM and AE.
Its \emph{memory interface} extracts vision-language features and combines them with retrieved history to condition the AE.
This pathway complements the original VLM-to-AE connection, which preserves direct access to the current scene. 

We distinguish \emph{slow parameters}, learned across episodes, from \emph{fast state}, updated within an episode.
Slow parameters include the policy, interface, memory projections, fusion gates, and fast-weight initialization $W_0$.
They learn how to encode and use history, which responsible for "\textit{how to memorize}"
Fast state $W_t$ consists of the current weights and biases of the memory networks and carries episode-specific information, which is "\textit{what to memorize}".
Tasks share slow parameters, while each episode starts from $W_0$ with an independent fast state.
During inference, all slow parameters remain fixed, while fast parameters undergo continuous self-supervised updates. During training, all parameters are unfrozen except those of the vision and language encoders.

\subsection{Observation-Grounded Memory Interface}
\label{sec:memory-interface}

The memory interface uses learned query tokens to aggregate VLM
features into a compact state summary for memory writes, retrieval,
and action conditioning (Fig.~\ref{fig:architecture}(b)).
This leverages pretrained vision-language representations without
updating memory over the full visual token sequence.
At depth $l$, interface representations $E_t^{(l)}$ extract
information from vision-language features $Z_t^{(l)}$:
\begin{equation}
  U_t^{(l)} = \operatorname{Attn}\!\left(
    Q_{\mathrm{if}}^{(l)}(E_t^{(l)}),
    K_{\mathrm{VL}}^{(l)}(Z_t^{(l)}),
    V_{\mathrm{VL}}^{(l)}(Z_t^{(l)})
  \right).
  \label{eq:interface-extraction}
\end{equation}
An attention mask restricts interface queries to valid vision-language
keys, excluding action, proprioceptive, and other interface tokens
while preserving the AE's original connections.
This grounds memory in observations and blocks a direct action-history
shortcut that could reduce imitation loss through action extrapolation
rather than task-state tracking.

The summary $U_t$ supports both retrieval and writing
(Fig.~\ref{fig:architecture}(c)).
Omitting layer/head indices, normalization, and positional encoding,
retrieval uses an observation-conditioned query:
\begin{equation}
  Q_t = U_t\theta_q,
  \qquad
  R_t = f_{W_t}(Q_t),
  \label{eq:interface-memory-read}
\end{equation}
where $f_{W_t}$ is the fast-weight memory and $R_t$ its
history-conditioned readout.
For writing, dedicated projections form key--value associations
from the same summary:
\begin{equation}
  K_t = U_t\theta_k,
  \qquad
  V_t = U_t\theta_v.
  \label{eq:interface-memory-write-projections}
\end{equation}
At scheduled write steps, a self-supervised update encodes these
associations into the fast weights:
\begin{equation}
  \begin{aligned}
    \mathcal{L}_{\mathrm{mem},t}
      &= \operatorname{mean}
         \left[(f_{W_t}(K_t)-V_t)^2\right], \\
    W_{t+1}
      &= W_t-\eta_t^{\mathrm{eff}}
         \nabla_{W_t}\mathcal{L}_{\mathrm{mem},t},
  \end{aligned}
  \label{eq:interface-memory-write}
\end{equation}
where $\eta_t^{\mathrm{eff}}=\eta_t\beta_t$ includes adaptive
scaling (Sec.~\ref{sec:adaptive-memory}).
Reads precede writes, so new observations affect subsequent decisions.
Memory projections $\theta_q$, $\theta_k$, and $\theta_v$ are separate
from the interface attention projections and learned through the outer
action objective; inner updates modify only episode-specific fast weights.

A channel-wise gate fuses retrieved history with the current
vision-language summary:
\begin{equation}
  \widetilde U_t^{(l)}
  = U_t^{(l)} + \tanh(\alpha^{(l)})\odot R_t^{(l)}.
  \label{eq:memory-fusion}
\end{equation}
Residual and feed-forward transformations then produce $C_t^{(l)}$.
Following RoboTTT~\citep{jiang2026robottt}, the gate initially limits
memory's contribution to protect pretrained visuomotor capabilities;
action supervision learns its channel-wise fusion strengths.

The enriched interface supplies keys and values to subsequent
action-token attention, alongside current vision-language features,
proprioception, and action context.
An interface update is consumed after its layer, not by that layer's
completed attention computation.
The AE thus combines VLM perception and retrieved history to predict actions.

\begin{figure}[t]
  \centering
  \includegraphics[width=\linewidth]{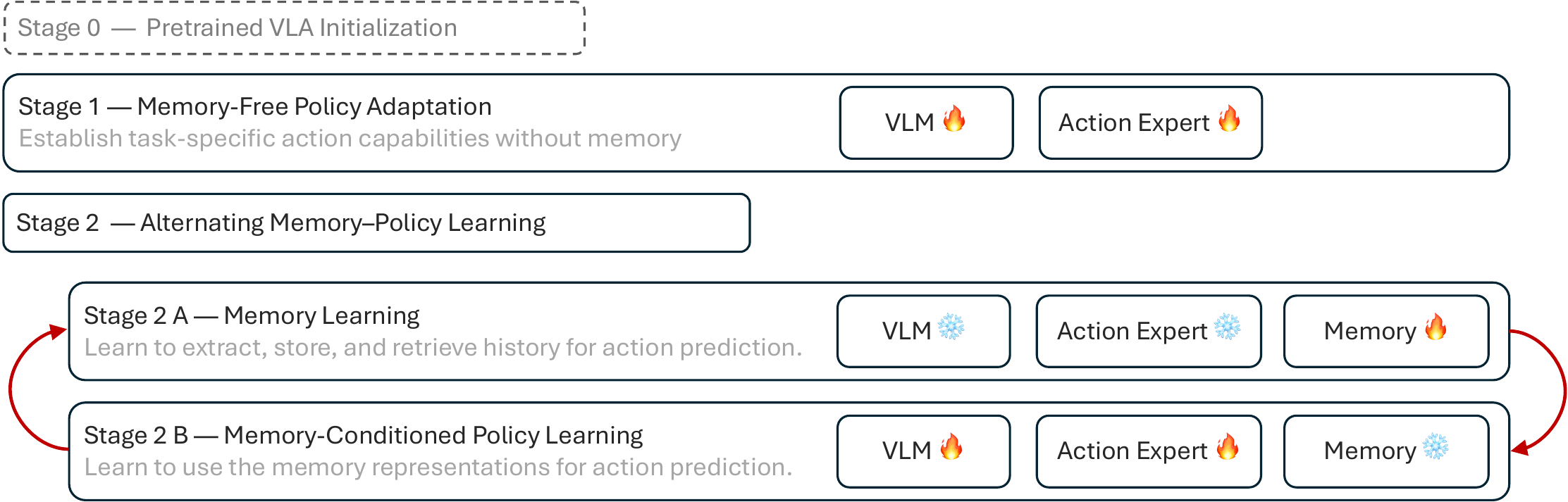}
  \caption{\textbf{Alternating memory--policy learning.}
  Stage 0 initializes a pretrained VLA, and Stage 1 adapts the
  policy without memory. Stage 2A learns memory with the policy
  fixed; Stage 2B learns a memory-conditioned policy with the
  memory mechanism fixed. Red arrows indicate repeated alternation;
  freezing refers to slow parameters, not episode-specific
  memory updates.}
  \label{fig:alternating-training}
\end{figure}

\subsection{Alternating Memory--Policy Learning}
\label{sec:alternating-learning}

Self-supervised association learning does not by itself ensure decision-relevant memory, and the policy must learn to use the new historical representations.
In joint training, memory updates change the context presented to the policy, while policy updates change the action gradients that guide memory learning.
The two modules may therefore continually adapt to each other's changing representations, making a stable memory-to-action mapping harder to learn.
We address this coupling by alternating their updates, holding one slow-parameter group fixed while optimizing the other.
Figure~\ref{fig:alternating-training} outlines the training stages, which is, \textbf{learning to remember and learning to act are separate.}

\textbf{Initialization and memory-free adaptation.}
We initialize from pretrained $\pi_{0.5}$ (Stage 0) and adapt
the policy without memory (Stage 1).
This establishes task-specific visuomotor skills before
introducing the memory pathway.

\textbf{Stage 2A: memory learning.}
We fix the VLM/AE and train the memory mechanism to extract,
store, and retrieve history useful for action prediction.
Expert action supervision passes through the frozen AE to shape
both current retrieval and earlier memory writes.
The fixed policy provides a stable decision-making counterpart,
encouraging memory representations that serve its control needs.

\textbf{Stage 2B: memory-conditioned policy learning.}
We fix the memory mechanism and adapt the policy to combine
current observations with retrieved history.
This phase adjusts both the vision-language features supplied
to memory and their use by the AE.
Only memory slow parameters are frozen: episode-specific fast
states continue to accumulate observations through online updates.

\textbf{Alternating schedule.}
We repeat these phases so that memory adapts to the current
policy, the policy learns to use the resulting representations,
and subsequent memory learning responds to the updated policy.
This repeated adaptation differs from training memory once
and then fitting a policy to it.
Both phases use expert action supervision without memory-content
or task-progress labels.
The action objective, parameter groups, sequence supervision,
and training schedule are detailed in Appendix~\ref{app:training}.

\subsection{Adaptive Self-Supervised Memory}
\label{sec:adaptive-memory}

Fast memory encodes history in an online-updated nonlinear mapping
(Fig.~\ref{fig:architecture}(c)).
The observation-grounded interface provides queries, keys, and values
for retrieval and self-supervised association learning.
This update requires no memory-content labels.
The initialization, projections, and step-size parameters are learned
through outer action supervision and fixed at deployment.

To limit repeated reinforcement from correlated observations, we adapt
write strength using the alignment between the descent direction and
$W_t-W_0$, together with the normalized reconstruction residual.
These signals determine a scale $\beta_t\in[0.1,1]$ that attenuates
aligned updates while preserving stronger updates for less aligned,
poorly reconstructed inputs:
\begin{equation}
  W_{t+1}=W_t-w_t\,\eta_t\,\beta_t\,
  \left.\nabla_W\mathcal L_{\mathrm{mem},t}(W)\right|_{W=W_t},
  \label{eq:adaptive-memory-update}
\end{equation}
where $w_t$ is the binary write mask and $\eta_t$ is a
curvature-calibrated step size with a learned positive multiplier.
The non-adaptive control fixes $\beta_t=1$ without changing
the step-size rule.

Reads precede writes, so each update affects only subsequent predictions.
The fixed-size fast state avoids a growing history buffer.
Architecture, scaling, and gradient details are provided in
Appendix~\ref{app:adaptive}; training and inference procedures
appear in Appendix~\ref{app:training}.

\section{Experiments}
\label{sec:experiments}

\label{sec:16table}

\begin{table*}[t]
\centering
\setlength{\tabcolsep}{4pt}
\renewcommand{\arraystretch}{1.12}
\resizebox{\textwidth}{!}{%
\begin{tabular}{ll*{17}{c}}
\toprule
\multicolumn{2}{l}{\multirow{2}{*}{\textbf{Method}}}
& \multicolumn{4}{c}{\textbf{Counting}}
& \multicolumn{4}{c}{\textbf{Permanence}}
& \multicolumn{4}{c}{\textbf{Reference}}
& \multicolumn{4}{c}{\textbf{Imitation}}
& \multirow{2}{*}{\textbf{AVG}} \\
\cmidrule(lr){3-6}\cmidrule(lr){7-10}\cmidrule(lr){11-14}\cmidrule(lr){15-18}
& & \makecell{Bin\\Fill} & \makecell{Pick\\Xtimes} & \makecell{Swing\\Xtimes} & \makecell{Stop\\Cube}
& \makecell{Video\\Umsk} & \makecell{Button\\Umsk} & \makecell{Video\\UmskS} & \makecell{Button\\UmskS}
& \makecell{Pick\\HighL} & \makecell{Video\\Repick} & \makecell{Video\\PlcBtn} & \makecell{Video\\PlcOrd}
& \makecell{Move\\Cube} & \makecell{Insert\\Peg} & \makecell{Pattern\\Lock} & \makecell{Route\\Stick} & \\
\midrule

\multicolumn{2}{l}{\textsc{Human Performance}} & 96.00 & 100.0 & 80.00 & 78.00 & 90.00 & 92.00 & 92.00 & 90.00 & 92.00 & 92.00 & 98.00 & 90.00 & 90.00 & 98.00 & 84.00 & 86.00 & 90.50 \\
\midrule
\multicolumn{19}{l}{\textbf{\textit{MME-VLA w/ Symbolic Memory}}} \\
\textsc{SimpleSG} & \texttt{Oracle} & 85.78 & 99.78 & 100.0 & 44.67 & 33.11 & 22.00 & 15.56 & 15.56 & 44.00 & 27.78 & 31.33 & 26.00 & 87.33 & 10.00 & 95.33 & 55.11 & 49.58 \\
\textsc{GroundSG} & \texttt{Oracle} & 85.78 & 100.0 & 100.0 & 49.67 & 98.78 & 95.00 & 99.22 & 80.22 & 83.33 & 97.33 & 100.0 & 100.0 & 87.78 & 15.56 & 97.00 & 55.56 & 84.08 \\
\addlinespace[3pt]
\multirow{2}{*}{\textsc{SimpleSG}} & \texttt{Gemini} & 46.00 & 63.00 & 45.00 & 2.00 & 29.00 & 9.00 & 14.00 & 2.00 & 20.00 & 15.00 & 26.00 & 29.00 & 61.00 & 4.00 & 7.00 & 0.00 & 23.25 \\
 & \texttt{QwenVL} & \textcolor{red!70!black}{\textbf{77.56}} & \textcolor{red!70!black}{\textbf{95.33}} & 5.11 & 0.44 & 34.22 & 19.33 & 15.33 & 9.56 & 17.11 & \third{25.33} & 33.33 & 25.11 & 82.00 & 3.78 & 12.67 & 7.78 & 29.00 \\
\multirow{2}{*}{\textsc{GroundSG}} & \texttt{Gemini} & 26.00 & 18.00 & 4.00 & 3.00 & 36.00 & 14.00 & 13.00 & 0.00 & 9.00 & 17.00 & 12.00 & 7.00 & 17.00 & 0.00 & 7.00 & 2.00 & 11.56 \\
 & \texttt{QwenVL} & 52.00 & \second{92.67} & 7.33 & 0.00 & \textcolor{red!70!black}{\textbf{88.67}} & 24.00 & \second{30.67} & 14.00 & 15.11 & \third{25.33} & \third{54.00} & \third{31.78} & 71.56 & 3.33 & 6.67 & 6.00 & 32.70 \\
\midrule
\multicolumn{19}{l}{\textbf{\textit{MME-VLA w/ Perceptual Memory}}} \\
\multirow{3}{*}{\textsc{TokenDrop}} & \texttt{Context} & 48.67 & 85.11 & \textcolor{red!70!black}{\textbf{94.67}} & 3.11 & 33.78 & 31.56 & 26.22 & 16.00 & 20.67 & 17.78 & 31.11 & 25.33 & 81.33 & 4.00 & 12.67 & 20.00 & 34.50 \\
 & \texttt{Modul} & 34.44 & 83.56 & 86.00 & 5.33 & 28.22 & 29.33 & 28.44 & \second{21.33} & 21.33 & 22.00 & \second{59.56} & \textcolor{red!70!black}{\textbf{36.00}} & 62.00 & \third{7.11} & \third{32.44} & \third{51.56} & 38.04 \\
 & \texttt{Expert} & 54.22 & 87.56 & \third{91.78} & 4.22 & 26.67 & 30.44 & 18.44 & 18.89 & 19.33 & 20.89 & 36.44 & 24.67 & \textcolor{red!70!black}{\textbf{87.56}} & 2.22 & 16.22 & 18.22 & 34.86 \\
\multirow{3}{*}{\textsc{FrameSamp}} & \texttt{Context} & 41.22 & 72.00 & 73.67 & 13.67 & 26.89 & 30.22 & 20.89 & 15.22 & 17.67 & 15.22 & 30.00 & 20.89 & 77.22 & 1.22 & 15.22 & 19.67 & 30.68 \\
 & \texttt{Modul} & 39.56 & 87.33 & \second{92.00} & \second{42.00} & 32.67 & 25.11 & 24.44 & 18.22 & \third{22.89} & \second{30.44} & \textcolor{red!70!black}{\textbf{60.00}} & \second{32.00} & 77.78 & \second{7.56} & \textcolor{red!70!black}{\textbf{53.56}} & \second{66.67} & \second{44.51} \\
 & \texttt{Expert} & \third{57.33} & 86.22 & \textcolor{red!70!black}{\textbf{94.67}} & \third{28.89} & 31.78 & 25.78 & 22.89 & \third{20.22} & 19.11 & 23.11 & 30.00 & 24.22 & \second{83.11} & 2.00 & 13.56 & 17.11 & 36.25 \\
\midrule
\multicolumn{19}{l}{\textbf{\textit{MME-VLA w/ Recurrent Memory}}} \\
\multirow{3}{*}{\textsc{TTT}} & \texttt{Context} & 35.56 & 62.89 & 42.44 & 3.33 & 29.78 & 22.89 & 18.44 & 14.44 & 20.44 & 13.11 & 34.22 & 20.22 & 32.44 & 1.11 & 1.56 & 3.56 & 22.28 \\
 & \texttt{Modul} & 34.22 & 65.11 & 36.67 & 2.11 & 27.22 & 22.11 & 25.22 & 14.11 & 14.56 & 12.11 & 32.67 & 22.33 & 31.22 & 1.11 & 3.56 & 7.00 & 21.96 \\
 & \texttt{Expert} & 34.89 & 63.78 & 41.33 & 4.00 & 31.78 & 22.44 & 19.56 & 18.00 & 12.22 & 9.56 & 34.00 & 22.89 & 33.56 & 0.89 & 3.11 & 5.56 & 22.35 \\
\multirow{3}{*}{\textsc{RMT}} & \texttt{Context} & 32.44 & 56.89 & 33.56 & 5.78 & 31.33 & 10.89 & 17.33 & 2.00 & 14.00 & 3.78 & 32.00 & 29.11 & 25.78 & 2.00 & 5.56 & 8.89 & 19.46 \\
 & \texttt{Modul} & 33.33 & 60.78 & 37.78 & 4.67 & 31.11 & 11.78 & 17.78 & 2.44 & 17.11 & 4.22 & 32.00 & 31.11 & 24.67 & 2.21 & 3.78 & 8.00 & 20.17 \\
 & \texttt{Expert} & 35.78 & 60.22 & 36.00 & 5.56 & 28.00 & 17.11 & 15.78 & 2.00 & 11.78 & 0.22 & 24.22 & 22.67 & 20.00 & 1.89 & 4.22 & 4.89 & 18.15 \\
\midrule
\multicolumn{19}{l}{\textbf{\textit{Other Methods}}} \\
\multicolumn{2}{l}{$\pi_{0.5}$} & 30.00 & 42.89 & 35.56 & 6.67 & 20.44 & 22.22 & 18.67 & 6.67 & 11.33 & 0.44 & 31.11 & 25.78 & 26.00 & 1.56 & 2.89 & 4.67 & 17.93 \\
\multicolumn{2}{l}{$\pi_{0.5}$ w/ past actions} & 26.67 & 58.33 & 26.67 & 4.67 & 30.67 & 23.67 & 20.67 & 16.00 & 12.33 & 8.67 & 24.00 & 18.67 & 34.00 & 1.00 & 4.00 & 5.67 & 19.73 \\
\multicolumn{2}{l}{SAM2Act+} & 40.00 & 76.00 & 25.33 & 0.00 & 27.33 & \third{32.00} & 18.00 & \textcolor{red!70!black}{\textbf{26.67}} & 17.33 & 5.33 & 24.67 & 20.00 & 29.33 & 0.00 & 0.00 & 0.00 & 21.37 \\
\multicolumn{2}{l}{MemER} & 56.67 & 79.33 & 59.33 & 0.00 & \third{81.33} & \textcolor{red!70!black}{\textbf{72.00}} & \textcolor{red!70!black}{\textbf{38.00}} & \second{21.33} & \textcolor{red!70!black}{\textbf{70.67}} & \third{25.33} & 30.00 & 26.00 & \third{82.67} & 6.67 & 16.67 & 12.00 & \third{42.38} \\
\midrule
\multicolumn{2}{l}{Stage 1(no memory)} & 26.67 & 34.00 & 31.33 & 2.00 & 27.33 & 6.00 & 18.00 & 2.00 & 11.33 & 1.33 & 26.67 & 27.33 & 38.00 & 1.33 & 3.33 & 5.33 & 16.38 \\
\multicolumn{2}{l}{\textbf{Ours}} & \second{60.67} & \third{91.33} & 90.67 & \textcolor{red!70!black}{\textbf{70.00}} & \second{88.00} & \second{59.33} & \third{29.33} & 20.00 & \second{50.67} & \textcolor{red!70!black}{\textbf{38.67}} & 35.33 & 30.00 & 80.67 & \textcolor{red!70!black}{\textbf{36.00}} & \second{50.00} & \textcolor{red!70!black}{\textbf{78.67}} & \textcolor{red!70!black}{\textbf{56.83}} \\[3pt]
\bottomrule
\end{tabular}%
}
\caption{Evaluation success rates (\%) on the 16 RoboMME tasks. \textcolor{red!70!black}{\textbf{Red}} denotes the highest score, \second{blue} the second-best, and \third{green} the third-best among non-privileged methods. Our method ranks among the top three on most tasks.}
\label{tab:main-results}
\end{table*}

Our experiments address four questions:
\textit{ (i) Can \tmem{} improve performance on memory-dependent tasks?
(ii) Which architectural and learning components contribute to its memory capability?
(iii) Does the policy rely on the content of its online memory?
(iv) How efficiently can the architecture perform inference?}
We first describe the evaluation setup, then examine task performance,
component ablations, online memory, and inference efficiency.

\subsection{Experimental Setup}
\label{sec:experimental-setup}

\textbf{Policy training.}
We initialize the policies from pretrained $\pi_{0.5}$~\citep{black2025pi05} and follow the training procedure in Sec.~\ref{sec:alternating-learning}.
We train on RoboMME data with 16 tasks.
Memory-free adaptation first establishes task-specific visuomotor skills.
We then alternate memory learning and memory-conditioned policy learning, using imitation supervision at valid execution frames in both phases.
Each trajectory starts from the learned fast-weight initialization, and its visual history updates memory independently of other trajectories.
Implementation details and the reference training recipe are given in Appendix~\ref{app:training}. The training takes 32 H200 for 48 hours.

\textbf{Benchmark and evaluation.}
RoboMME contains 16 manipulation tasks in four suites: Counting, Permanence, Reference, and Imitation~\citep{dai2026robomme}.
These probe temporal, spatial, object-centric, and procedural memory.
Our standard evaluation uses the complete 50-episode test set with three evaluation seeds. The policy predicts 64 actions and executes 32 before replanning.
The ablation and counterfactual experiments in Secs.~\ref{sec:ablation-studies} and~\ref{sec:online-memory} use task-specific checkpoints.

\textbf{Baselines.}
All baseline scores in Table~\ref{tab:main-results} are quoted directly from the original RoboMME paper~\citep{dai2026robomme}.
They include symbolic memory (SimpleSG and GroundSG), perceptual memory (TokenDrop and FrameSamp), and recurrent memory (TTT and RMT).
Additional references are $\pi_{0.5}$, $\pi_{0.5}$ with past actions, SAM2Act+, and MemER~\citep{dai2026robomme,fang2025sam2act,sridhar2026scaling}.
Human and privileged Oracle results are reference points rather than deployable competitors.
Baseline provenance and differences in information access are distinguished in Appendix~\ref{app:evaluation}.

\subsection{Performance on Memory-Dependent Tasks}
\label{sec:memory-task-performance}

We first examine whether \tmem{} enables effective control on memory-dependent tasks.
Table~\ref{tab:main-results} summarizes the RoboMME evaluation (including no memory baseline).
Among methods without privileged information, \tmem{} ranks within the top three on most reported tasks, with the highest success rates on VideoRepick, InsertPeg, and RouteStick.
It also substantially improves over the memory-free $\pi_{0.5}$ and the evaluated recurrent-memory baselines.

Performance is strongest on tasks that primarily require direct retrieval
of an earlier cue or tracking repeated events, such as VideoUnmask and
StopCube. In contrast, success remains lower on VideoUnmaskSwap and
ButtonUnmaskSwap, where the policy must track swaps and update
object--location associations rather than simply recall a stored binding.
This contrast suggests a distinction between retaining information and
reasoning over it. We hypothesize that, as a single-model approach, \tmem{} relies on the
underlying $\pi_{0.5}$ policy, whose per-frame
representations are not trained for cross-frame correspondence, for visual feature extraction and temporal
reasoning, without an external reasoning model. Its memory supplies
historical evidence but does not, by itself, confer the ability to infer
how that evidence changes through subsequent events. The weaker swap
performance may therefore reflect limitations in the base policy($\pi_{0.5}$)'s
temporal reasoning, which improved memory retention alone cannot resolve. Designing implicit memory modules that can track and reason over evolving task states is left to future work.

Low-level control imposes a separate limitation on tasks such as InsertPeg.
Our qualitative observations on InsertPeg indicate that the policy can identify and approach the intended target yet fail during the final insertion.
Thus, terminal success reflects both memory-dependent decision-making and execution precision.
The strong relative improvement on InsertPeg, despite its modest absolute success rate, highlights the importance of distinguishing these failure sources.

\subsection{Ablation Studies}
\label{sec:ablation-studies}

We examine three components of \tmem{}: the memory interface,
alternating memory--policy learning, and adaptive memory writing. Two tasks are used to do the experiments: VideoUnmask (Appendix Figure \ref{fig:task-videounmask})and Movecube(Appendix Figure \ref{fig:task-movecube}). The model of the ablation experiment was trained independently on each individual task. Both variants use identical single-task data, batch size, and initialization.

\begin{figure*}[t]
  \centering
  \includegraphics[width=0.94\textwidth]{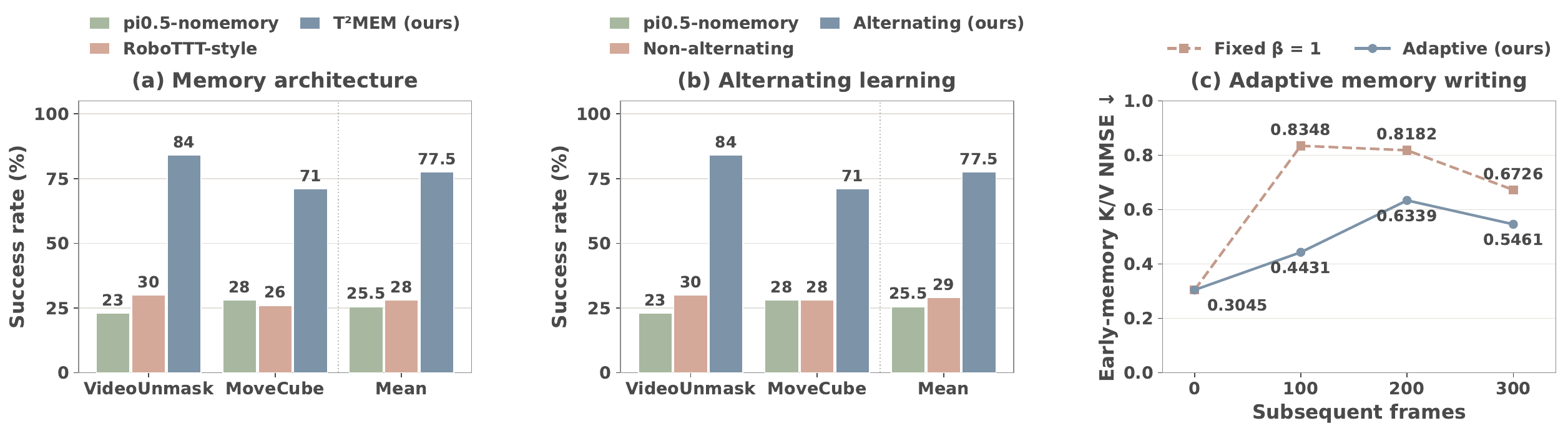}
  \caption{\textbf{Ablation studies of \tmem{}.}
  (a) Memory architecture and (b) training schedule: success rates on
  VideoUnmask and MoveCube, with a memory-free $\pi_{0.5}$ reference.
  Mean denotes the average across the two tasks.
  (c) Adaptive writing: reconstruction NMSE of early-memory K/V bindings
  over subsequent observations, averaged over 20 VideoUnmask Hard episodes.
  Lower is better; the label $\beta=1$ denotes $\beta_t=1$ at every write, disabling adaptive scaling but retaining step-size calibration.
  Training configurations and comparison scope are detailed in Appendix~\ref{app:interventions}.}
  \label{fig:ablations}
\end{figure*}

\textbf{Memory architecture.}
We compare our architecture with a RoboTTT
integration~\cite{jiang2026robottt} on VideoUnmask and MoveCube.
The latter places memory within the action expert and derives memory
reads and writes from action-expert representations. Figure~\ref{fig:ablations}(a) compares these architectures
with a memory-free policy. \tmem{} achieves 84\% and 71\% success on the
two tasks, compared with 30\% and 26\% for the AE-side implementation. 
Appendix~\ref{app:interventions} describes the training configurations. This suggests that, unlike retaining past actions for long-context processing, memory-dependent tasks require semantic features extracted from the VLM. T²MEM supports this through its observation-grounded memory interface.

\textbf{Alternating memory--policy learning.}
We evaluate alternating and non-alternating optimization on VideoUnmask and
MoveCube. Non-alternating training updates memory and policy parameters jointly.
Alternating optimization repeatedly adapts memory representations
to the policy and the policy to those representations.
Figure~\ref{fig:ablations}(b) reports joint-training results. In same training budget, non-alternating training shows results similar to no-memory baseline.

\textbf{Adaptive memory writing.}
We compare adaptive writing with a $\beta_t=1$ control on 20 long
VideoUnmask Hard episodes, keeping the architecture, write cadence,
and step-size normalization unchanged. Starting from shared
visible-demonstration memory, we replay 300 subsequent real frames
and measure reconstruction error on the initial K/V bindings.
Figure~\ref{fig:ablations}(c) shows 18.8\%--46.9\% lower error
with adaptive writing, consistently across all 20 episodes.
This supports reduced interference with earlier memory bindings
under real-trajectory replay.

\subsection{Contribution of Online Memory}
\label{sec:online-memory}

We intervene at inference time to test whether task success depends on online writes, stored content, and observation timing.
These diagnostics use fixed task-specific checkpoints for VideoUnmask, MoveCube, and SwingXtimes, with intervention-specific write schedules detailed in Appendix~\ref{app:interventions}.

\begin{figure}[t]
    \centering
    \includegraphics[width=0.94\linewidth]{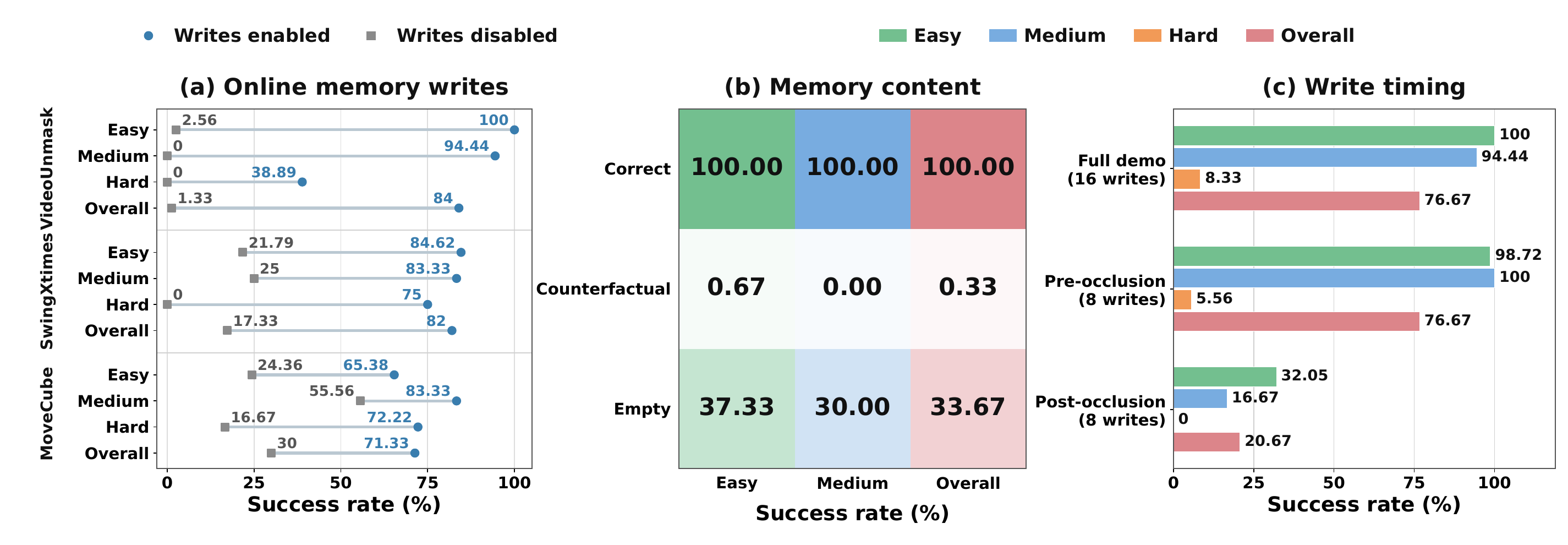}
    \caption{\textbf{Online memory supports history-dependent decisions.}
    (a) Disabling fast-weight writes reduces success across three tasks.
    (b) Correct, conflicting, and empty memories produce markedly different outcomes on VideoUnmask.
    (c) Pre-occlusion writes match full-demonstration performance with half the updates, whereas post-occlusion writes perform substantially worse.
    Success rates (\%) are pooled over three seeds under each panel's evaluation protocol.}
    \label{fig:online memory}
\end{figure}

\textbf{Disabling online writes.}
We disable episode-local fast-weight updates while retaining memory reads and the learned initialization.
Each condition covers all 50 test episodes and three seeds.
Success drops by 82.67 percentage points on VideoUnmask, 64.67 on SwingXtimes, and 41.33 on MoveCube (Figure~\ref{fig:online memory}a).
The fixed policy and initial memory alone therefore cannot sustain normal performance, demonstrating the importance of online updates, instead learned a genearal solution to all tasks.

\textbf{Replacing memory content.}
We construct 50 counterfactual pairs with conflicting targets from VideoUnmask.
We reconstruct each target episode, retaining its instruction, environment goal, and diffusion seed, while supplying correct demonstration memory, conflicting donor memory, or empty memory (Appendix~\ref{app:interventions} and Figure~\ref{fig:conter}).
Across both pairing directions and three seeds, these conditions yield 300/300, 1/300, and 101/300 successes, respectively (Figure~\ref{fig:online memory}b).
Conflicting memory is more damaging than empty memory, supporting content-specific use of history.
Empty-memory performance is close to the one-third chance level for three candidates, which is close random selection in no memory policy.

\textbf{Localizing useful writes.}
We restrict VideoUnmask writes to different demonstration windows, using stride 4 to provide sufficient updates within each window.
The 66-frame prefix contains 16 write opportunities; an image audit identifies the dominant occlusion transition at frame 32 in all eight inspected trajectories.
We compare all 16 writes with eight pre-occlusion or eight post-occlusion writes.
Under equal write budgets, pre-occlusion writing achieves 76.67\%, versus 20.67\% after occlusion, and matches full-demonstration performance (Figure~\ref{fig:online memory}c).
These results localize decision-relevant evidence to the interval in which the target remains visible.

\subsection{Inference Efficiency}
\label{sec:inference-efficiency}

\begin{figure}[t]
    \centering
    \includegraphics[width=0.75\linewidth]{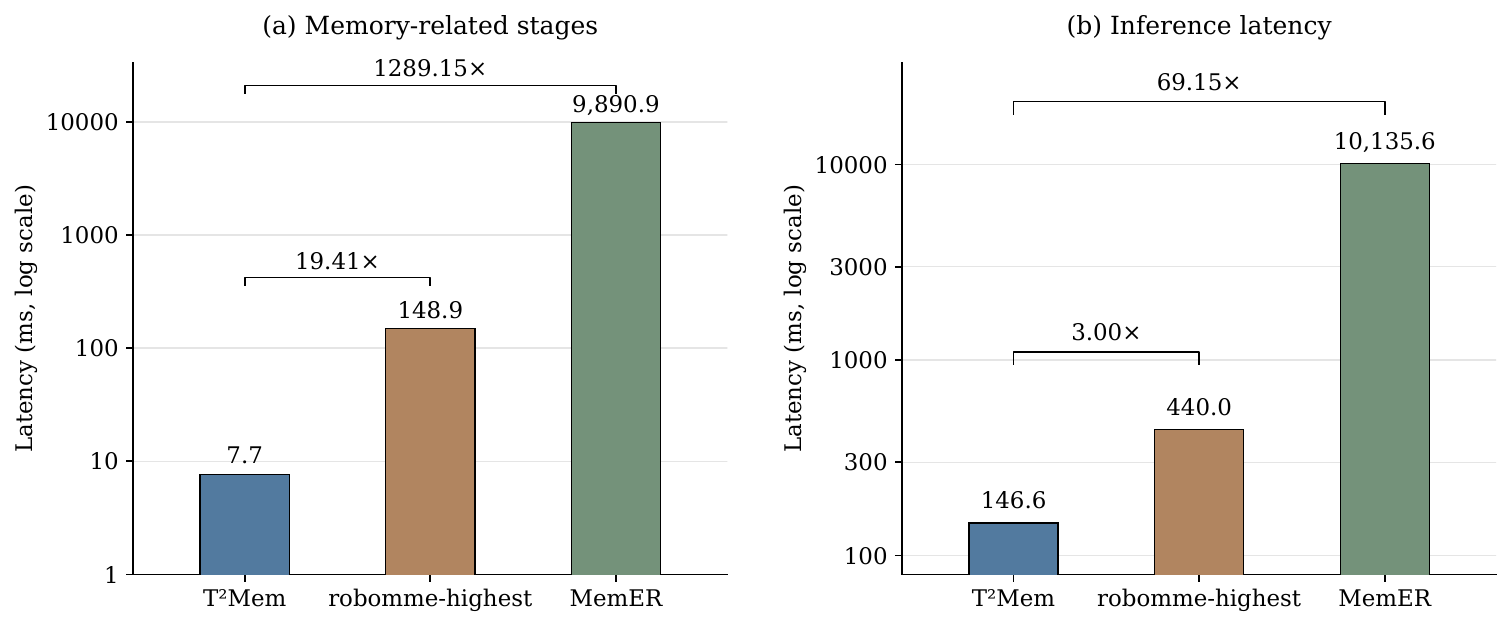}
    \caption{\textbf{Controlled-workload inference efficiency.}
    Left: latency introduced by memory only.
    Right: end to end latency for an inference.}
    \label{fig:inference-efficiency}
\end{figure}

We profile \tmem{}, FrameSamp+Modul in RoboMME, and MemER on RTX A5000 GPUs at batch size one to test the efficiency.
Timing uses device synchronization after warm-up, excluding initialization, compilation, communication, and environment execution. As shown in Figure~\ref{fig:inference-efficiency}, our foreground computation is approximately $3.0\times$ faster than FrameSamp+Modul and $69.2\times$ faster than MemER.
Profiling boundaries and workload configurations are detailed in Appendix~\ref{app:profiling}. These savings reflect a single-model design that stores and retrieves history in latent space, without external autoregressive reasoning.
Its compact parametric memory supports low-latency decisions from historical cues, facilitating fast closed-loop operation on memory-dependent tasks.

\section{Conclusion}
\label{sec:conclusion}
We presented \tmem{}, a framework that makes memory an internal
capability of a robot policy. An observation-grounded interface
connects vision-language perception to fast-weight memory, whose
online self-supervised updates retain history for action prediction.
Alternating memory--policy learning decouples learning to remember
with learning to use memory through expert action supervision,
without memory-specific annotations or external model calls.
Experiments on the 16 RoboMME tasks demonstrate improved
memory-dependent manipulation, while interventions show that
decisions depend on both online updates and the information stored.
These findings support
learning memory formation and use within a single policy as
a practical direction for history-dependent robotic control. The remaining challenges lie in the foundation model's intrinsic capabilities for cross-temporal reasoning and information extraction, which will be explored in the future. 

\bibliography{main}
\bibliographystyle{plainnat}

\clearpage
\appendix
\section{Architecture and Adaptive Memory Updates}
\label{app:details}
\label{app:architecture}

\paragraph{Interface and layer correspondence.}
The VLM and action expert each contain 18
Transformer layers. Sixteen learned interface tokens extract vision--language
information and carry memory readouts toward the action expert. They are
updated through attention, gated memory fusion, and feed-forward blocks,
without reinitialization between layers. The fast state persists across
observations within an episode. Memory modules are allocated at all 18 layers; because each interface update is consumed only by subsequent layers, 17 of them are effective. Parameter counts below include all 18.

\begin{table}[t]
\centering\small
\begin{tabular}{@{}p{0.30\linewidth}p{0.64\linewidth}@{}}
\toprule
Component & Reference configuration \\
\midrule
VLM / action expert & 18 layers each; widths 2,048 / 1,024 \\
Interface & 16 tokens of width 1,024, initialized from $\mathcal N(0,0.02^2)$; rank-16 attention adapters \\
Fast-weight memory & 16 heads; per-head MLP $64\rightarrow256\rightarrow64$, exact GeLU, biases \\
Learned initialization & $W_0$: matrix standard deviation 0.02, zero biases \\
Read/write projections & Separate biased $1{,}024\rightarrow1{,}024$ Q/K/V projections \\
Normalization / position & Q/K/V RMS normalization ($\epsilon=10^{-6}$); Q/K interleaved RoPE, base $10^4$ \\
Fusion / state precision & Channel-wise $\tanh(\alpha)$ initialized to 0.01; FP32 fast state \\
\bottomrule
\end{tabular}
\caption{Core architecture settings. RoPE uses interface token positions, not elapsed episode time.}
\label{tab:app-architecture}
\end{table}

\paragraph{Attention and parameterization.}
VLM tokens cannot attend to the policy suffix. Interface queries attend only
to valid VLM keys, not to interface, proprioceptive, or action tokens.
Proprioceptive and action queries retain native suffix attention, including
access to the interface. The direct current-observation pathway is preserved.
The fusion gate controls memory's contribution to pretrained features; it is
distinct from the adaptive write scale below.

The approximately 67.90M added parameters comprise 56.678M read/write
projections, 9.529M learned fast-state initialization parameters, 1.622M
interface attention adapters, and approximately 0.069M interface embeddings,
gates, step multipliers, and proprioceptive projection parameters. An episode's allocated fast state contains
9,529,344 FP32 scalars (36.35 MiB), excluding attention caches and activations. 

\subsection{Adaptive Write Rule}
\label{app:adaptive}

For one layer, let $K_t,V_t$ be observation-derived bindings and $W_t$ the
pre-write state. With $H_m$ memory heads, $N$ interface tokens, and head width $d$,
\begin{equation}
 \mathcal L_t=\frac{1}{H_mNd}\|f_{W_t}(K_t)-V_t\|_F^2,
 \qquad g_t=\nabla_{W_t}\mathcal L_t.
\end{equation}
The following quantities are computed per head, with the head index omitted.
Set $\Delta_t=W_t-W_0$ and $u_t=-g_t$. Define
\begin{equation}
 a_t=\frac{\langle u_t,\Delta_t\rangle}{\max(\|u_t\|\|\Delta_t\|,10^{-12})},
 \quad r_t=\operatorname{clip}_{[0,1]}
 \frac{\operatorname{RMS}(f_{W_t}(K_t)-V_t)}{\max(\operatorname{RMS}(V_t),10^{-8})}.
\end{equation}
With $n_t=1-\operatorname{clip}_{[0,1]}(a_t)$ and $z_t=n_t(1+r_t)/2$,
\begin{equation}
 \beta_t=0.1+0.9(3z_t^2-2z_t^3),\qquad W_{t+1}=W_t-\beta_t\eta_t g_t.
 \label{eq:app-write}
\end{equation}
For squared direction norm at most $10^{-24}$, set $n_t=0$; otherwise, for
squared displacement norm at most $10^{-24}$, set $n_t=1$. This rule modulates
writing using reconstruction residuals and accumulated changes. 

\paragraph{Curvature calibration.}
Let $G=\|g_t\|^2$ and $C=2\|J_fg_t\|^2/(H_mNd)$, with sums restricted to the
current head. Using its contribution $\mathcal L_h$ to the total loss,
\begin{align}
 q&=\operatorname{clip}_{[10^{-12},10^3]}\left(G/\max(C,10^{-30})\right),\\
 b&=\frac{\mathcal L_h(W_t-qg_t)-\mathcal L_h(W_t)+Gq}{q^2},\\
 \ell_t&=\min\!\left(10^3,
 \begin{cases}G/\max(2b,10^{-30}),&b>0,\\q,&b\leq0,\end{cases}\right),
 \quad \eta_t=0.5\,\operatorname{softplus}(\rho)\ell_t.
\end{align}
Each layer learns one $\rho$, initialized to $\log(e-1)$, giving an initial
multiplier of one. The cap applies to $\ell_t$, not the final learned product.
There is no additional inner gradient clipping; zero gradient gives no change.
Both $\beta_t$ and curvature calibration are stop-gradient quantities. Gradients
remain through the fast-state recurrence, $g_t$, projections, and learned
multiplier, rather than replacing the entire update with a first-order
approximation. The non-adaptive control in Figure~\ref{fig:ablations}(c)
sets $\beta_t=1$ at every write: it retains curvature calibration and does not imply constant
$\eta_t$.

\section{Training Algorithm and Hyperparameters}
\label{app:training}

\paragraph{Shared action objective.}
Let $\phi$ denote memory slow parameters and $\theta$ the trainable policy parameters.
For a valid execution frame and its expert action chunk $A_t$, we construct
$x_t^\tau=\tau\epsilon+(1-\tau)A_t$ at valid positions and minimize
\begin{equation}
  \mathcal L_{\mathrm{act}}(\theta,\phi)
  = \mathbb E\!\left[
    \left\|v_{\theta,\phi}(x_t^\tau,o_t,\ell,W_t,\tau)
    -(\epsilon-A_t)\right\|_{M_t}^{2}
  \right].
  \label{eq:method-action-loss}
\end{equation}
Here, $o_t=(I_t,p_t)$, $\epsilon$ is Gaussian noise, and $\tau$ is flow time.
The mask $M_t$ selects valid future actions and supervised dimensions;
each chunk's loss is normalized by its number of valid elements.
The fast state $W_t$ is constructed recursively from preceding observations,
allowing action gradients to supervise current retrieval and earlier writes
through the retained sequence.
The inner objective $\mathcal L_{\mathrm{mem}}$ updates fast state from
observations, whereas the outer objective $\mathcal L_{\mathrm{act}}$ learns
the slow parameters that form and use that state.
Both alternating phases use the same action objective, without memory-content
or task-progress labels.

\paragraph{Stage-specific parameter groups.}
Stage 1 adapts VLM LoRA and the AE, including its action input/output and time
projections, on individual execution frames and future action chunks without memory.
Stage 2A fixes $\theta$ and updates $\phi$: the interface and its low-rank
adapters, memory projections, fast-weight initialization, inner step-size
parameters, and fusion gates. Gradients pass through the frozen AE into memory.
Stage 2B fixes $\phi$ and updates VLM LoRA, the full AE, action input/output
and time projections, and the proprioceptive projection. VLM LoRA can change
the features supplied to the fixed memory mechanism; fast-state reads and
writes remain active. The visual encoder and base VLM weights stay frozen
throughout training.

\paragraph{Reference recipe.}
Table~\ref{tab:app-training} records the training process of 16 tasks.

\begin{table}[t]
\centering\small
\begin{tabular}{@{}p{0.18\linewidth}p{0.24\linewidth}p{0.24\linewidth}p{0.24\linewidth}@{}}
\toprule
 & Stage 1 & Stage 2A & Stage 2B \\
\midrule
Purpose & Memory-free adaptation & Memory learning & Memory-conditioned policy learning \\
Initialization & Pretrained $\pi_{0.5}$ & Stage-1 step 20,000; new memory & Preceding memory phase \\
Updated group & VLM LoRA, complete AE, action input/output and time projections & Interface, memory projections and initialization, step multipliers, gates & VLM LoRA, AE, action/time and proprioceptive projections \\
Budget & step 20,000 & 500 updates per cycle & 500 updates per cycle \\
Learning rate & $2\!\times\!10^{-5}$ & $7.5\!\times\!10^{-5}$; gate $5\!\times\!10^{-4}$ & $3\!\times\!10^{-5}$ \\
Effective batch & 64 frames (8 $\times$ accumulation 8) & 32 sequences & 32 sequences \\
\bottomrule
\end{tabular}
\caption{Released training recipe.}
\label{tab:app-training}
\end{table}

\begin{figure}[t]
\begin{minipage}{\linewidth}\small
\textbf{Training: one alternating-phase update}
\begin{enumerate}\setlength{\itemsep}{2pt}
\item Select the active slow-parameter group and a task-homogeneous batch.
\item Initialize independent fast states $W\leftarrow W_0$ and loss $L\leftarrow0$.
\item Traverse observations chronologically. Read memory and accumulate masked
action loss at valid execution frames.
\item At scheduled writes, update memory from the observation using
Eq.~\ref{eq:app-write}. Demonstration frames may write but have no action loss.
\item Backpropagate through the sequence; update only the active slow group
and its optimizer state.
\end{enumerate}
\medskip
\textbf{Inference: one episode}
\begin{enumerate}\setlength{\itemsep}{2pt}
\item Reset $W\leftarrow W_0$; process any demonstration prefix chronologically.
\item At replanning, hold $W$ fixed throughout action denoising. Predict 64
actions using repeated memory reads.
\item Commit one observation-conditioned write if scheduled. Do not write
once per denoising iteration.
\item Execute 32 actions, process the midpoint observation for its stride-16
write, and replan at the next chunk boundary.
\end{enumerate}
\end{minipage}
\caption{Training and inference order. Reads precede writes at the same observation.}
\label{fig:app-algorithms}
\end{figure}

\paragraph{Time units.}
Stride 16, prediction horizon 64, and execution horizon 32 use physical
environment steps, not optimizer updates or denoising iterations. The midpoint
execution observation can enter memory without replanning; other intermediate
frames do not each cause a write. Demonstration sampling uses its own indexed
observation timeline. Stride-4 diagnostics are specified separately
in Appendix~\ref{app:interventions}.

\section{Evaluation Protocols and Baseline Configurations}
\label{app:evaluation}

The main evaluation uses 50 RoboMME test episodes per task: 26 easy, 12 medium,
and 12 hard. The policy predicts 64 actions and executes 32 before replanning.
Success follows the benchmark's terminal predicate; partial progress is not
counted. Scores pool 150 rollouts per task over three evaluation seeds and
are macro-averaged across the 16 tasks.

\paragraph{Symbolic memory.}
SimpleSG represents history through language subgoals; GroundSG additionally
specifies target locations in front-view image coordinates. Subgoals are
appended to the task instruction for the $\pi_{0.5}$ policy. They are supplied
by prompted Gemini-2.5-Pro, Qwen3-VL-4B fine-tuned on subgoal annotations,
or simulator ground truth (Oracle). The learned predictor uses the current
image and previous subgoals, rather than retaining a visual history buffer.
Oracle results therefore involve privileged information.

\paragraph{Perceptual and recurrent memory.}
TokenDrop retains visual patches selected by temporal RGB differences,
whereas FrameSamp uniformly samples historical frames.
TTT compresses visual history into fast weights through self-supervised
updates; RMT recurrently updates learned memory tokens.
Each neural representation is evaluated with three integration mechanisms:
\emph{Context} appends memory tokens to the VLM input;
\emph{Modul} uses memory cross-attention to produce adaptive LayerNorm
conditioning for the action expert; and \emph{Expert} introduces a separate
memory transformer attended to by the action expert.

\paragraph{Additional reference methods.}
The memory-free $\pi_{0.5}$ uses the current observation, while its
past-action variant appends action history to the language tokens.
SAM2Act+ uses a SAM2-based memory bank and predicts discrete keyframe
waypoints executed by the simulator's motion planner.
MemER combines stored keyframe images with VLM-generated subgoals.
RoboMME adapts MemER by fine-tuning Qwen3-VL-4B with grounded-subgoal
and keyframe annotations, then executing its predictions with GroundSG.
Unlike SimpleSG and GroundSG, its subgoal predictor receives accumulated
visual evidence, not only the current image and subgoal history.

\section{Memory Intervention and Ablation Details}
\label{app:interventions}

\paragraph{Fixed diagnostic policies.}
Online-memory interventions use VideoUnmask step 2,000 (stride 4), SwingXtimes
step 2,000 (stride 16), and MoveCube step 2,000 (stride 16), all with H64/E32. Write
suppression and timing interventions evaluate all 50 test episodes per seed, as shown in Table \ref{tab:app-interventions}

\begin{table}[t]
\centering\small
\begin{tabular}{@{}p{0.20\linewidth}p{0.72\linewidth}@{}}
\toprule
Intervention & Protocol \\
\midrule
Disable writing & Zero effective updates from reset, including demonstration
and execution. Retain $W_0$, reads, and normal inference-call/RNG cadence. \\
Replace content & Build memory from the target demonstration, a conflicting
donor demonstration, or no demonstration. Keep the target environment,
instruction, and sampling seed fixed across conditions. Empty means learned
$W_0$, not all-zero weights. \\
Restrict write times & On a 66-frame prefix, permit all $\{2,6,\ldots,62\}$
writes, only $\{2,6,\ldots,30\}$, or only $\{34,38,\ldots,62\}$. The write-timing protocol restricts writes to a 66-frame prefix, which truncates longer Hard demonstrations; the full-prefix condition therefore differs from the unrestricted setting in panel (a).\\
\bottomrule
\end{tabular}
\caption{Protocols for the three panels of Figure~\ref{fig:online memory}.}
\label{tab:app-interventions}
\end{table}

\paragraph{Content and timing controls.}
The content diagnostic uses 50 pairs (25 easy, 25 medium, generated, not in training). Both directions and
three seeds give $50\times2\times3=300$ evaluations per condition. We evaluate the same
target episodes with correct, conflicting, or empty demonstration memory.
For the timing comparison, we vary the permitted demonstration writes around
frame 32, the occlusion transition identified in eight inspected trajectories.
The pre- and post-occlusion conditions each permit eight demonstration writes.

\paragraph{Adaptive-writing diagnostic.}
The retention probe uses 20 VideoUnmask
episodes. Eight prefix
observations at frames $2,6,\ldots,30$ build a common adaptive state. Both
conditions clone it and retain the original K/V bindings for measurement.
Real observations at $34,38,\ldots,330$ then produce 25, 50, and 75 writes at
the reported 100-, 200-, and 300-frame offsets.

For each saved frame and layer,
\begin{equation}
 \operatorname{NMSE}=
 \frac{\operatorname{mean}_{h,n,d}(f_W(K)-V)^2}
 {\max(\operatorname{mean}_{h,n,d}V^2,10^{-12})}.
\end{equation}
Ratios are averaged equally over saved frames and allocated memory layers,
then over episodes. Only $\beta_t$ differs between conditions; curvature calibration
and write cadence are unchanged. Aggregate paired-win counts report lower adaptive
NMSE in all 20 episodes at each nonzero offset.

\paragraph{Architecture and training controls.}
The AE-side comparison uses memory over action-expert representations as RoboTTT in \cite{jiang2026robottt}, whereas
\tmem{} uses its VL interface.
The AE-side results in Figure~\ref{fig:ablations}(a) use step-2,000 checkpoints. Training alternates
500 memory and 500 policy updates, giving 1,000 updates to each group at this
checkpoint. Evaluation uses H64/E32 with stride-16 writes, including midpoint
execution observations. Apart from the architecture, all other settings—including alternating training, adaptive writing, and the number of training steps—are identical in architecture ablation.

The non-alternating baseline jointly updates all trainable memory, VL-adapter, and policy parameters at every step. All other settings match the single-task alternating run: initialization from Stage-1 step 20,000 with fresh memory, and per-task official demonstrations. Memory, gate, and policy learning rates are $6\times10^{-5}$, 
$5\times10^{-4}$, and $3\times10^{-5}$, with 25-step warm-up.
The reported joint-training results use step 2,000, with H64/E32.

\begin{figure*}[t]
  \centering
  \includegraphics[width=\textwidth]{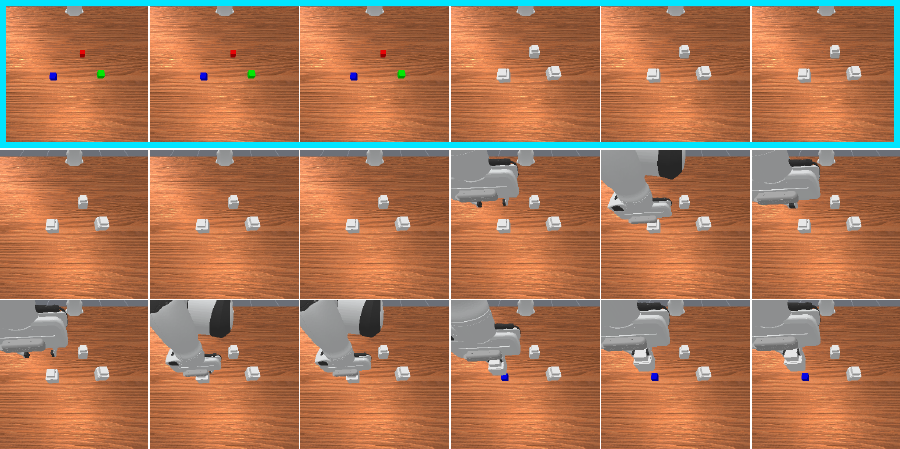}
  \caption{\textbf{VideoUnmask: recalling information after occlusion.}
  The robot observes colored cubes, then lifts the container hiding the
  instructed color. In this training trajectory, color--location bindings
  are visible in frames $0$--$31$ and occluded from frame $32$ (the 33rd frame).
  Choosing the correct container therefore requires recalling the early cue.
  This localized evidence window enables targeted interventions on memory
  acquisition. Selected frames follow temporal order from left to right,
  then top to bottom. The blue border marks the video demonstration, during which
the policy observes without executing actions.}
  \label{fig:task-videounmask}
\end{figure*}

\begin{figure*}[t]
  \centering
  \includegraphics[width=\textwidth]{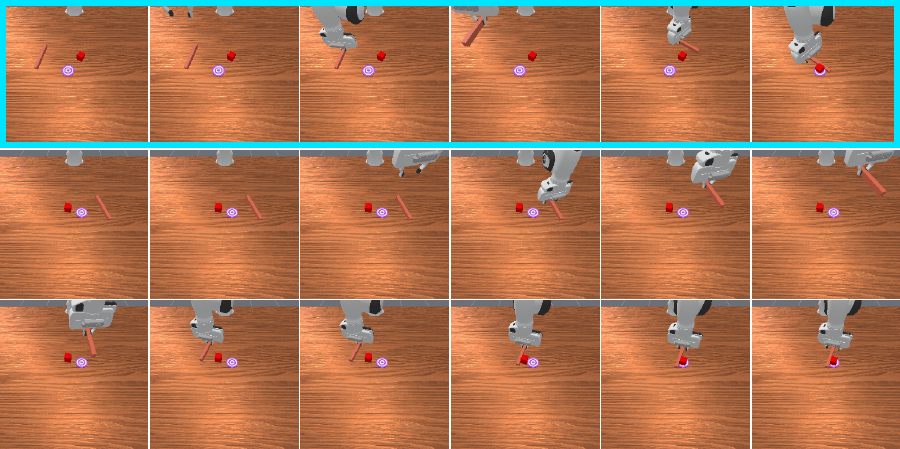}
  \caption{\textbf{MoveCube: remembering how an action was demonstrated.}
  The robot watches a demonstration and moves a cube to the target using the
  same method: pushing, pick-and-place, or tool-mediated hooking.
  This training example shows hooking in the demonstration and subsequent
  execution. Because the instruction does not specify the method, the policy
  must remember how the cube was moved, not merely its destination.
  Frames are ordered from left to right, then top to bottom. The blue border marks the video demonstration, during which
the policy observes without executing actions.}
  \label{fig:task-movecube}
\end{figure*}

\begin{figure*}[t]
  \centering
  \includegraphics[width=\textwidth]{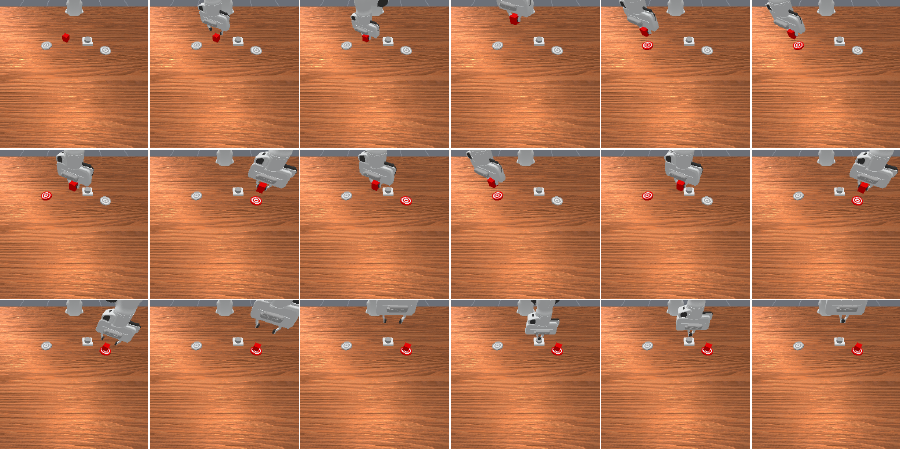}
  \caption{\textbf{SwingXtimes: remembering progress through repeated actions.}
  The robot moves the red cube between two targets for two right-to-left
  cycles, then puts it down and presses the stop button.
  Similar configurations recur across cycles, so the policy must track
  completed visits to stop after the instructed count.
  Frames are ordered from left to right, then top to bottom;
  left and right follow the robot's coordinate frame.}
  \label{fig:task-swingxtimes}
\end{figure*}

\begin{figure}
    \centering
    \includegraphics[width=1\linewidth]{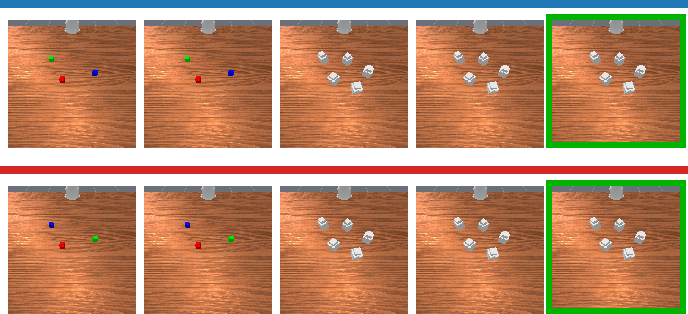}
    \caption{Counterfactual histories in VideoUnmask. The two rows show different color assignments before occlusion but identical observations afterward. Given the same instruction to select the container hiding the green cube, the correct choice depends on the observed history rather than the current image alone.}
    \label{fig:conter}
\end{figure}

\section{Inference Profiling}
\label{app:profiling}

\paragraph{Setup.}
We profile steady-state calls on RTX A5000 GPUs at batch size one, with device
synchronization after warm-up. \tmem{} and FrameSamp+Modul use three warm-up
calls and 30 measurements; MemER's high-level model uses three and 20, and
its low-level policy uses four and 30. Initialization, compilation, RPC,
simulation, and action execution are excluded.

\begin{figure}[!t]
\centering
\includegraphics[width=0.8\linewidth]{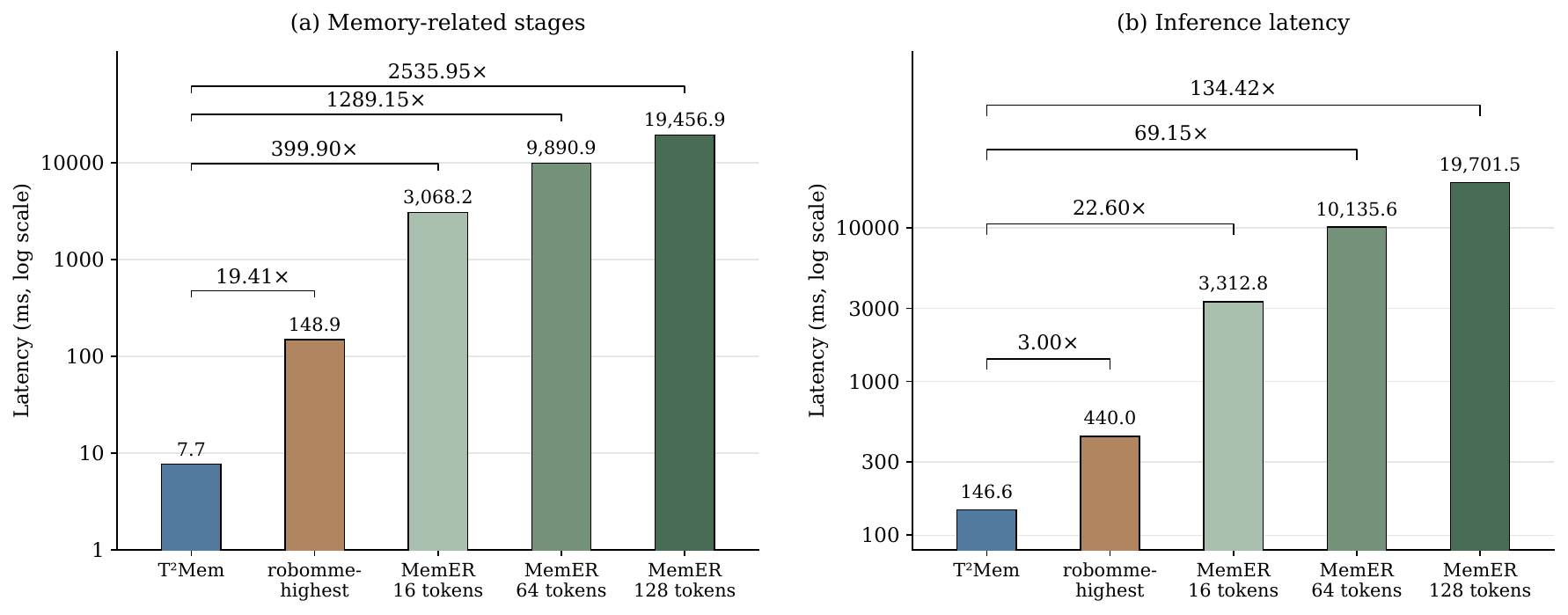}
\caption{Sensitivity to the external planner's generation budget.}
\label{fig:app-decode-sensitivity}
\end{figure}

FrameSamp+Modul
uses 512 memory tokens (32 selected images). MemER uses
Qwen3-VL-4B-Instruct, rank-16 LoRA, BF16 and SDPA, with eight recent images,
eight retained keyframes, and 64 generated tokens. Comparasion between different token buget in MemEr is in Figure \ref{fig:app-decode-sensitivity}.

\end{document}